\documentclass{IEEEtran} 
\usepackage{amsmath}

\usepackage{hyperref}

\usepackage[utf8]{inputenc} 
\usepackage[T1]{fontenc} 
\usepackage{listings} 
\usepackage{graphicx}

\usepackage{listings}

\newcommand{\ie}{{\it i.e. }} 
 
\DeclareMathOperator\erf{erf}

\newcommand{\tp}{\tilde{P}}

\newlength{\singlefigwidth}
\newlength{\doublefigwidth}
\hypersetup { pdftitle = {Title}, pdfsubject = {Subject}, pdfauthor = {Piotr Bialas} }
\begin{document} \lstset{language=Python}

\title{A multi-instance deep neural network classifier: application to Higgs boson CP measurement.} 
\author{Piotr~Bialas and Elzbieta~Richter-Was \\
  Jagiellonian University, ul. Łojasiewicza 11, 30-348 Krakow, Poland\\
  Daniel Nemeth \\
  Eötvös Loránd University, 1117, Budapest, Pázmány Péter sétány 1/A, Hungary}
 
\maketitle
\begin{abstract}
	We investigate properties of a  classifier applied 
	to the measurements of the CP state of the Higgs boson in $H\rightarrow\tau\tau$ decays.
        The problem is framed as binary classifier applied to individual instances. Then the
        prior knowledge that the instances belong to the same class is used to define the
        multi-instance classifier. Its final score is calculated as multiplication of single
        instance scores for a given series of instances. In the paper we discuss properties of such classifier, notably its dependence on the number of instances in the series. This classifier exhibits very strong random dependence on the number of epochs used for training and requires
		careful tuning of the classification threshold. We  derive formula for this optimal threshold.   
\end{abstract}

\section{Introduction}

The Deep Neural Networks (DNN) \cite{LeCun2015} have been shown to work very well across many different domains, including image classification,
machine translation or speech recognition. Recently, it is also finding his place in the applications to very demanding classification
problems in High Energy Physics (HEP) \cite{HEPNN_1, HEPNN_2}. 

In this paper we present further development of the DNN application reported in \cite{Jozefowicz:2016kvz}, where the possibility of measuring
the CP state of the Higgs boson produced in pp collisions at LHC accelerator,  using a DNN trained on the Monte-Carlo data was investigated.
The problem was defined as
binary classification problem with the goal to distinguish between two different CP states of the Higgs boson.
Solution presented in \cite{Jozefowicz:2016kvz}  was concentrated on quantifying  performance of a single instance classifier
\ie on predicting the probability that a given instance is a CP scalar object with the alternative hypothesis of being pseudo-scalar or mixed state.
As a measure of the performance an  area under the receiver operational characteristic curve (AUC)~\cite{roc} was used.

What was not explicitly explored in \cite{Jozefowicz:2016kvz} is a prior knowledge that only one of the different CP states can be
realized in the nature according to the considered model, so the sample of multiple instances will belong to the one class only.
This discussion will be a subject of the work presented here. 

Classification is probably one of the most common machine learning (ML) tasks. One possible approach is to use Bayesian classifier.
This amounts to calculating or estimating for each category $C_i$ the conditional probability 
$P(C_i| X)$ that, given the input variables (features) $X$,  analyzed instance belongs to the category $C_i$.

In this paper we will consider only the case of binary classifiers, with two categories denoted as $A$ and $B$. Classification then consists of
comparing  probability $P(A|X)$ with some threshold $\theta$.  We will call the classifier using $P(A|X)$  a {\em single-instance} classifier.
In practice, often one cannot find such set of features and $\theta$ which cleanly separate the two categories.

There are however cases when one can assume that a sample of $N$ instances, denoted by $\{X_i\}$, consists of all $X_i$
belonging to the same category. We can then use this information to greatly increase the accuracy of classification.
This can be achieved by calculating the probability  $P(A|\{X_i\})$ that given the features $\{X_i\}$ all instances in
a sample belong to category $A$.  We will call the classifier  estimating  $P(A|\{X_i\})$  a {\em multi-instance} classifier.

In this paper we will discuss how to calculate properties of the  {\em multi-instance} classifier, notably its dependence on the sample size $N$,
from the properties of the single instance classifier. We will propose also how to choose optimal threshold  to assure that predictions of {\em multi-instance} classifier are regularised, i.e. are not too sensitive to the number of epochs used
for training DNN. 

\section{Data}

Without going into details on the nature of the problem and its practical importance, lets us briefly remind that we are
discussing measurement of the properties of the Higgs boson, recently discovered by the experiments at CERN LHC proton-proton
collider. This resonance searched for since decades by HEP experiments, is an evidence of the mechanism explaining
within the context of so called Standard Model how elementary particles are acquiring their masses.
The presently available statistics of the Higgs boson samples allows to explore  ML techniques to measure quite precisely its internal
properties, like the spin and CP state, crucial to support that indeed observed resonance is a Higgs boson of the Standard Model.
The case studied here is quite challenging,
the Higgs boson is decaying into two objects (tau leptons),  $H\rightarrow\tau\tau$, each of them decaying further into objects
which caries in the correlations between they directions information about the CP state of the initial resonance we are interested in.
So the goal of the DNN algorithm will be to identify  those correlations in the multi-dimensional phase-space and use them
 for classifying the instances as belonging to one or other category allowed by the model.

For the case studied here we use the same Monte Carlo data as studied in ~\cite{Jozefowicz:2016kvz}
but we consider only one classification case, namely the $H\rightarrow\tau\tau\rightarrow a_1^\pm\rho^\mp2\nu$ decay,
with  $a_1\pm\to 3 \pi^\pm$ and $\rho^\mp \to \pi^\pm \pi^0$. The problem is defined as classification based on  $7\times4 = 28$ input variables in total,  namely
7 outgoing $\tau$ - pair decay products, each represented by 4-vector in the energy-momenta phase-space. We do not build any functional
features out of those variables, but we use directly 4-vector representation, however in the frame which, after boost and rotation from the
laboratory frame, removes trivial symmetries from the system which then do not have to be rediscovered by the DNN.
This choice of the frame was discussed in details in~\cite{Jozefowicz:2016kvz} and is motivated by the nature of the problem.
One should note that those variables are not independent because of the kinematic constraints.
Some of them are also not detected experimentally, so the set of $X_i$ being 28 variables for each instance $i$ represents
the idealistic scenario.  We will call this set a {\em complete} data set.
We also consider a more realistic scenario, including only momenta of the particles which can be detected experimentally,
i.e. removing neutrinos $\nu$ from the list. This gives in total $5\times 4=20$ variables for each instance $i$.
This more realistic scenario was used also in~\cite{Jozefowicz:2016kvz}. We will call this set an {\em incomplete} data set. 

Because in this paper we are interested in the methodology and properties of the classifiers rather then physics problem
itself, as a case study we consider discrimination between two possible scenarios: a CP scalar which we will denote as $A$ vs.
mixed scalar-pseudoscalar state with mixing angle $\phi^{CP}=0.4$ denoted as $B$. 

The available statistics of Monte-Carlo (MC) data is approximately four millions of instances. Each instance, defined by
the momenta of all the decay products, has two weights associated with it denoted by $\omega_A$ and $\omega_B$. Those weights
are respectively proportional to the probability that a given instance is of class $A$ or $B$ respectively. Those weights
are calculated by the Monte Carlo program used for simulating physics model of interest and are depending
on  $7\times4 = 28$ input variables used in {\em complete} data set.
In the case of {\em incomplete} data set the weights $\omega_A$ and $\omega_B$ are still calculated using $7\times4 = 28$ variables,
but as inputs to DNN is missing some of those, weights are no longer representing an unique function of the inputs.

\section{Binary classification}

When  conditional probability $P(A|X)$ is available the classification depends on a single threshold parameter $\theta$.
We classify that an instance belongs to category $A$ when
\begin{equation}
	P(A|X)>\theta.
\end{equation}
and to category $B$ otherwise.

In practice one cannot usually find such set of features and $\theta$ which cleanly separate the two categories and as a
consequence one is faced with misclassification errors. Those errors are usually quantified by two metrices: true positives
rate (TPR) and false positives rate (FPR)
\begin{equation}
	\begin{split}
		TPR(\theta) =\frac{\text{number of }A\text{ classified as }A}{\text{number of }A}\\
		FPR(\theta) =\frac{\text{number of }B\text{ classified as }A}{\text{number of }B}
	\end{split}
\end{equation}
The TPR and FPR values depend on the threshold parameter $\theta$. If we consider $(FPR(\theta), TPR(\theta))$ as a point on the plane,
varying $\theta$ will follow a curve known as Receiver Operational Characteristic (ROC) curve~\cite{roc}. 

If the probability distributions are known, the TPR and FPR can be calculated as follows 
\begin{equation}
	\begin{split}
		TPR(\theta) =\int\text{d}X P(X|A) \Theta(P(A|X)-\theta)\\
		FPR(\theta) =\int\text{d}X P(X|B) \Theta(P(A|X)-\theta)
	\end{split}
\end{equation}
where $\Theta$ denotes the Heaviside function equal to zero if its argument is negative and one otherwise.
The $P(X|A)$ is the probability density  for $X$ variables in category $A$ and similarly for $P(X|B)$. 

The area under ROC curve (AUC score) is another measure of the quality of the classifier. It is equal to the  probability
that a positive (A) instance will be rated higher then a negative (B) instance~\cite{roc}
\begin{equation}\label{eq:AUC}
	\begin{split}
		AUC = \int & \text{d}X^A P(X^A|A)\text{d}X^B P(X^B|B) \\
		&\Theta(P(A|X^A) - P(A| X^B)).
	\end{split}
\end{equation}
The value of AUC score equal to one half corresponds to random classification and value of $AUC = 1$ indicates a {\em perfect} classifier.

\section{Single instance classifier}

 We started by training a DNN  with  inputs $X$ and outputs $w$ defined as 
\begin{equation}
	w(X)=\{w_1(X), w_2(X)\}=\frac{\{\omega_A(X), \omega_B(X)\}}{\omega_A(X) + \omega_B(X)} 
\end{equation}
The probability that instance $X$  belongs to category $A$   is just  $w_1$ 
\begin{equation}
	P(A|X)=\frac{\omega_A(X)}{\omega_A(X)+\omega_B(X)} = w_1(X).
\end{equation}
We use cross-entropy for the loss function
\begin{equation}
	LOSS = \frac{1}{M}\sum_{i=1}^M\sum_{j=1}^2 w_j(X_i)\log\tilde{w_j}(X_i) 
\end{equation}
where $M$ is the number of instances used for training. The tilde over a variable denotes DNN output.

For training we use the Keras\cite{Keras} framework with the TensorFlow~\cite{TensorFlow} backend. We use seven layers DNN
with softmax activation on the last layer. Remaining layers use the PReLU rectifier~\cite{prelu} as well as batch normalization~\cite{batch}.
This model is similar to the one that was proposed in~\cite{Jozefowicz:2016kvz}.
The code defining the model and implementation is shown in the Listing~\ref{lst:model}. 
\begin{lstlisting}[float,caption = Python code defining the Keras model., captionpos=b, language=Python, tabsize=4, label=lst:model, frame=lines] 
layer_size = 320 
dropout_rate =0.18 
kmodel = Sequential()
	
kmodel.add(Dense(input_dim=num_features, 
           units=layer_size)) 
kmodel.add(BatchNormalization()) 
kmodel.add(PReLU()) 
kmodel.add(Dropout(dropout_rate)) 
for i in range(5): 
	kmodel.add(Dense(units=layer_size)) 
	kmodel.add(BatchNormalization()) 
	kmodel.add(PReLU()) 
	kmodel.add(Dropout(dropout_rate)) 
	
kmodel.add(Dense(units=2)) 
kmodel.add(Activation('softmax'))
	
kmodel.compile(
       loss='categorical_crossentropy', 
       optimizer='Adam', metrics=['mse']) 
\end{lstlisting}

\begin{figure}
	\begin{center}
		\includegraphics[width=\doublefigwidth]{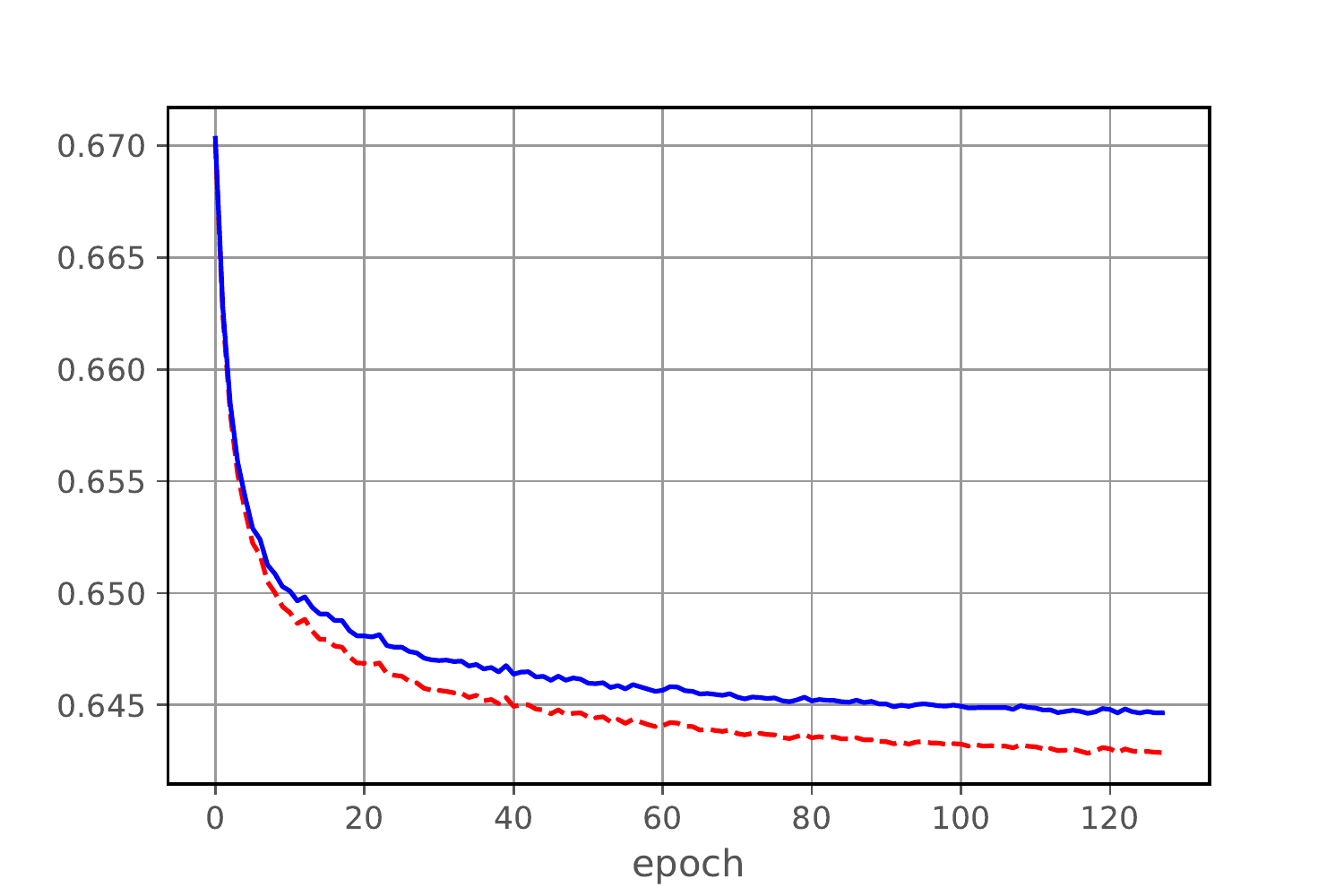}\\
		\includegraphics[width=\doublefigwidth]{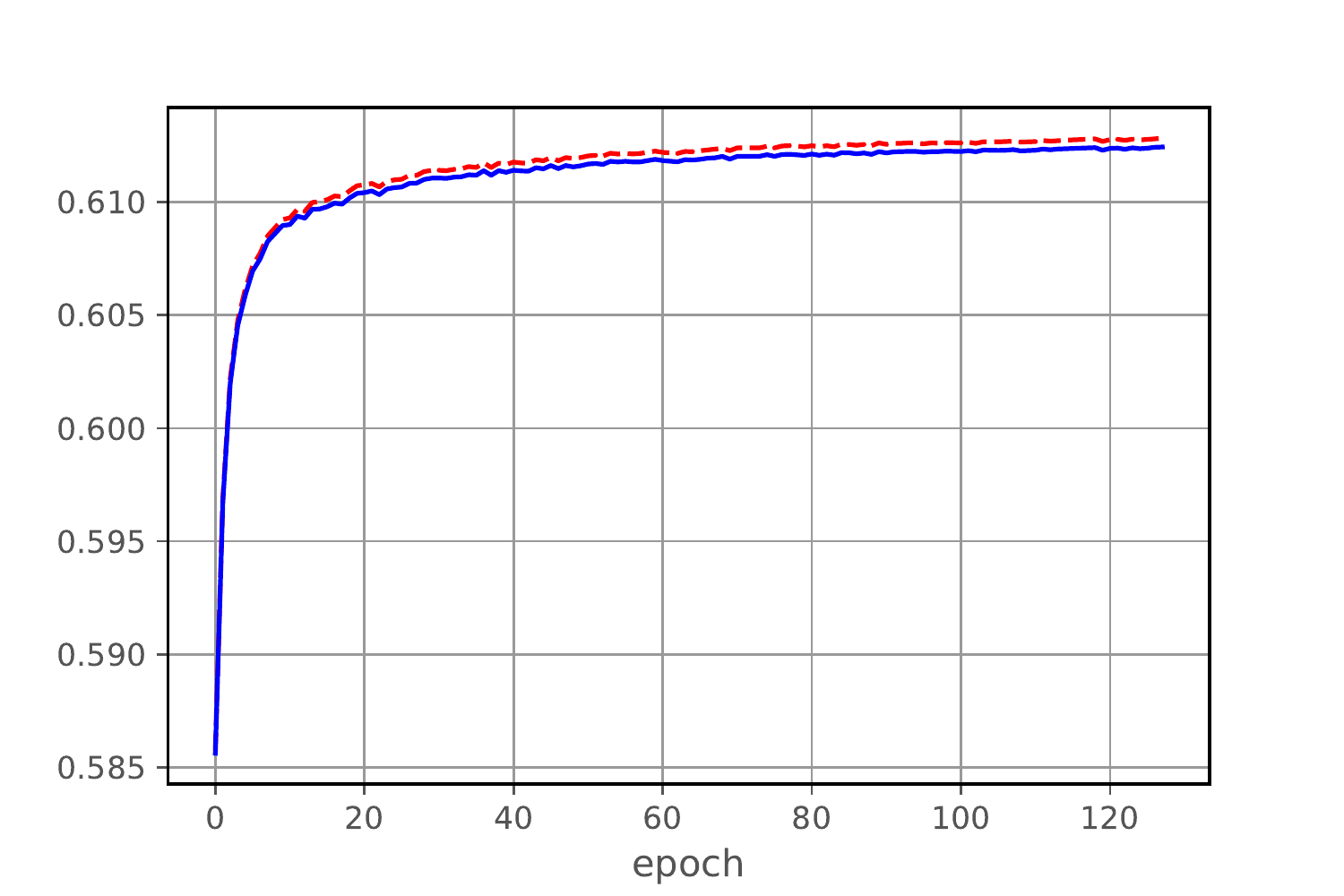} 
		\caption{\label{fig:only-mom-loss}The LOSS (top) and AUC score (bottom) as a function of number of epochd used for training
                  the network on {\it complete} data set.
                  Dashed line (red) denotes the training and solid line (blue) the validation results. Dropout rate used $5\%$. } 
	\end{center}
\end{figure}

Let us discuss first {\it complete} data set. In this case the outputs $w$ are a function (in mathematical sense) of the inputs \ie outputs are
uniquely determined by the inputs. The best possible LOSS and AUC score  can be calculated directly from
the MC data sample using approximation
\begin{equation}\label{eq:approx-int}
	\int\text{d}X P(X|A) f(X) \approx \frac{\sum_{i=1}^M w^{A}_i f(X_i)}{\sum_{i=1}^M w^{A}_i}.
\end{equation}
and similarly for $B$, yielding  $0.615$ for the AUC score. This is consistent with what reported in Table 2 (top-right column) of~\cite{Jozefowicz:2016kvz}
for the same data set and equivalent case.
The evolution of the LOSS and AUC score as function of number of epochs used for the neural-network training and with a $5\%$ dropout is shown
in the Figure~\ref{fig:only-mom-loss}. As can be seen from those plots the network gets close to the best possible AUC score (0.615) and
does not overfit. 

Lets us move now  {\it incomplete} data set. In this case the mapping from $X$ to outputs $w$ is not unique \ie not a
function in the mathematical sense. We tried the same network architecture and  dropout, but the network significantly overfitted.
This likely is due to the fact that the data in this case looks like much more noisy as the same set of input features $X$ potentially
describes several different instances and so corresponds to different values of $w_i$.
To fix this issue we increased dropout rate and after some trials we have settled on the dropout of $18\%$. 
The achieved performance scores are much lower than for {\it complete} data, with the plateau  AUC score of  about $0.535$ (instead of  $0.615$)
but it is to be expected. The nature of the problem is that in the {\it incomplete} data set we removed some features (4-vectors)
which are essential for picking up on correlations which allow the discrimination between model $A$ and $B$.
DNN cannot  learn the original mapping as the $w_i$ outputs are no longer unique function of the $X_i$ variables and the network
is performing some form of averaging.
In this case we cannot calculate maximal possible AUC score directly from the data as we have no practical means to estimate  $P(A|X_i)$. 

\begin{figure}
	\begin{center}
		\includegraphics[width=\doublefigwidth]{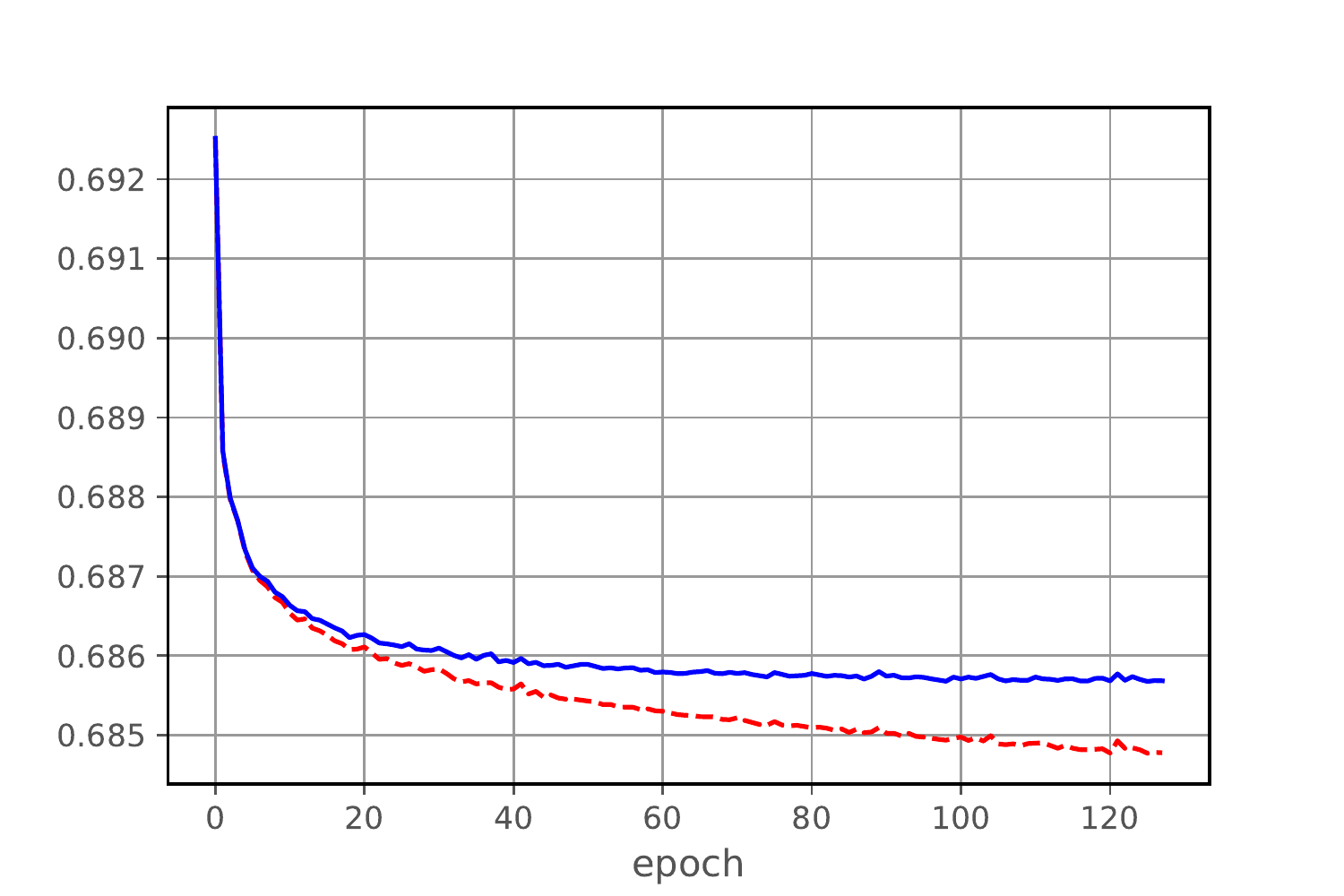}\\
		\includegraphics[width=\doublefigwidth]{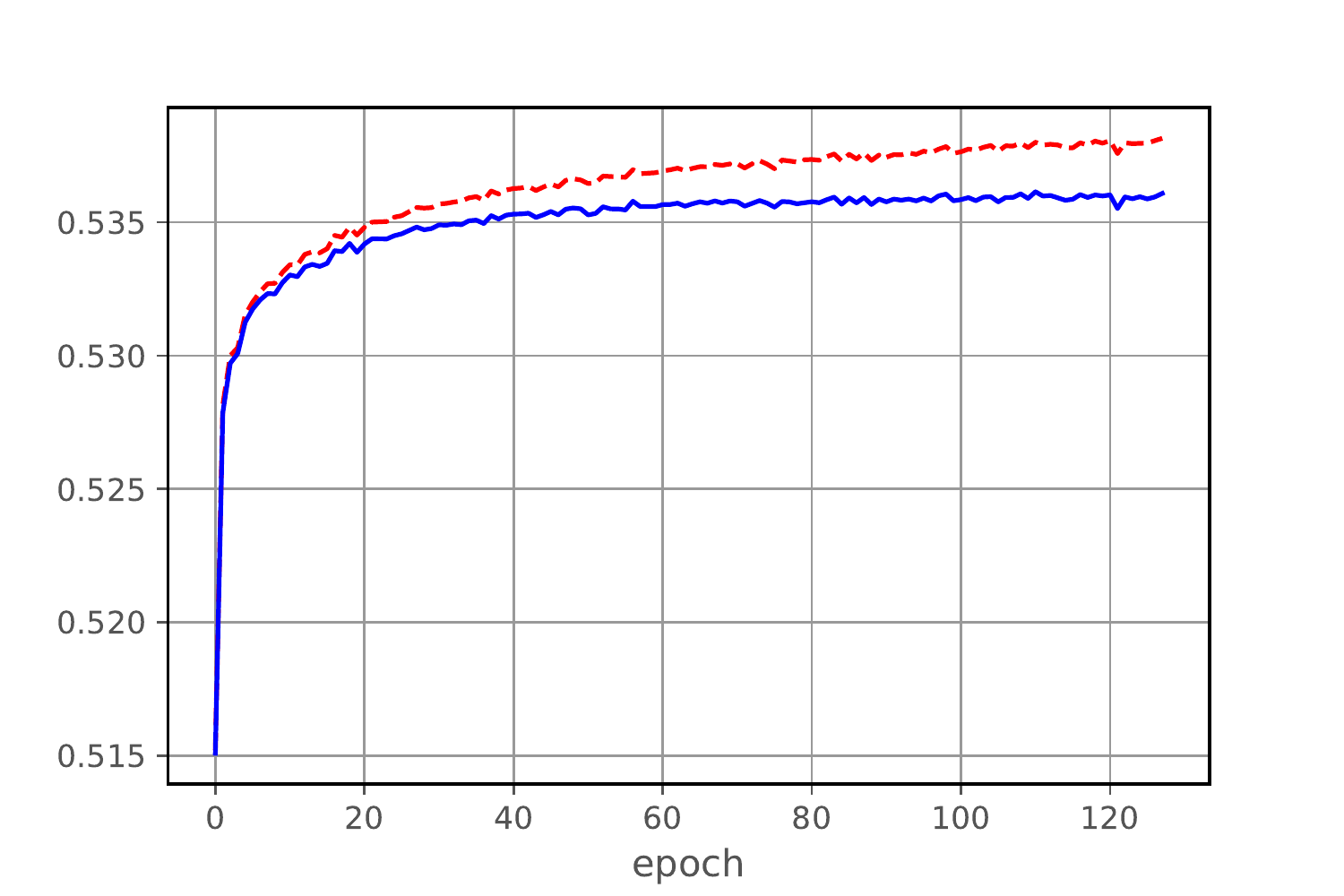}
		\caption{\label{fig:only-all-non-loss}The LOSS (top) and AUC score (bottom) for the network trained on {\it incomplete} data set.
                  Dashed line (red) denotes the training and solid line (blue) the validation results. Dropout rate used $18\%$.} 
	\end{center}
\end{figure}

\begin{figure}
	\begin{center}
		\includegraphics[width=\singlefigwidth]{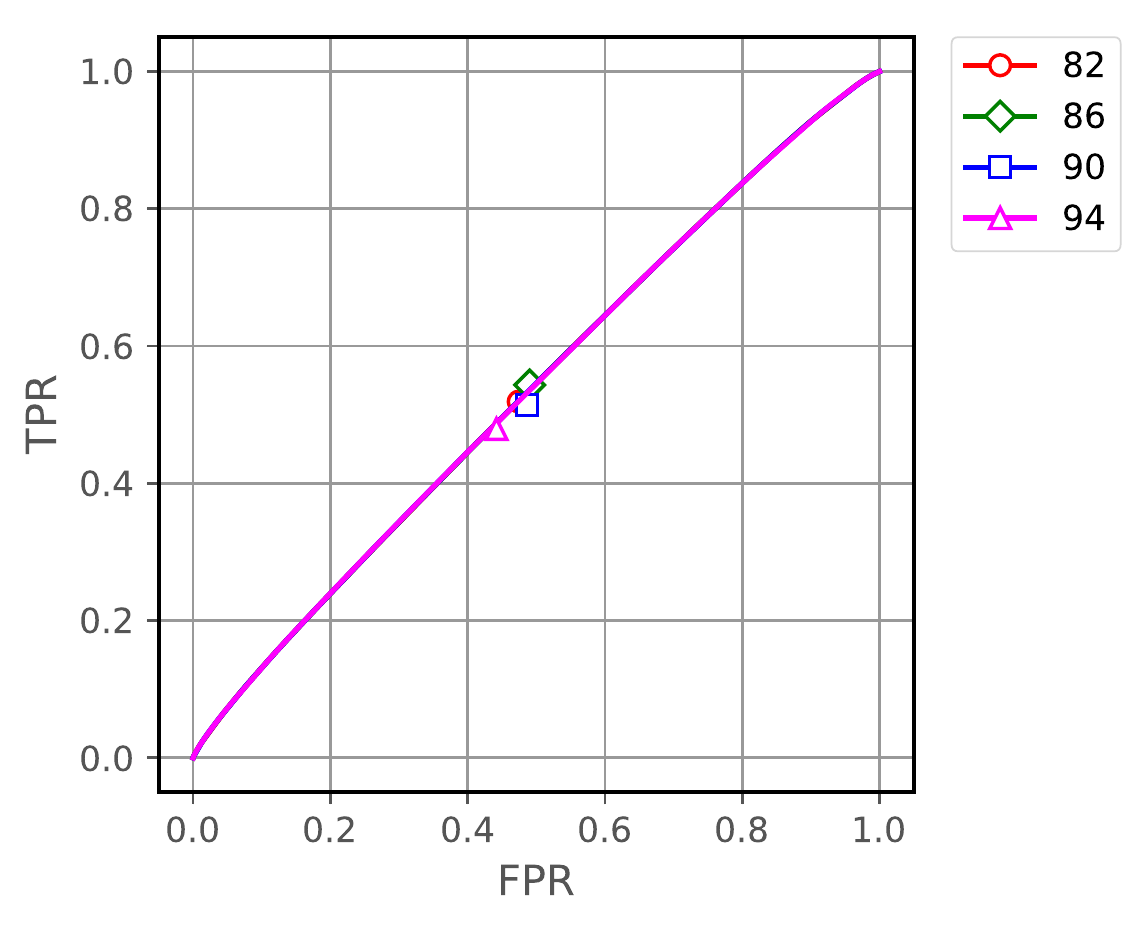}%
	\end{center}
	\caption{\label{fig:multi-roc}ROC curves for  networks  after different number of epochs. The points correspond to the $\theta=0.5$
          threshold.} 
\end{figure}

So far we have considered only the AUC score as a quantitative metric of the classifier performance. In  practice we might be more interested
in the actual true and false positives rates.
In Figure~\ref{fig:multi-roc} we show the corresponding ROC curves for four arbitrarily chosen number of epochs
used for training DNN. 
As we can see all curves fall on top of each other which is consistent with the fact that the AUC score does not change much between epochs
after some plateau is reached, as shown in the Figure~\ref{fig:only-all-non-loss} (bottom).
We have also marked in the same plot the true and false positive rates
corresponding to threshold $\theta=0.5$ for each of those networks and they do not coincidence. 
This indicates that the classification threshold $\theta$ should be adjusted further, individually for each network (labeled by number of epochs
used for training) to make predictions less sensitive to small variations due to the number of epochs used in the training.
In the next section we will discuss how to calculate the optimal threshold $\theta$.  

\section{Multiple instance classification}

As discussed in the previous section a single instance classifier cannot reliably distinguish between two different categories,
the best AUC score being of  about $0.535$ for single instance and  the {\em incomplete} data set only. 
However, as we are discussing the problem were all instances must belong to the same category we can increase the accuracy of the classification
by simultaneously interpreting results of classification of $N$ sequential instances.
We will discuss this below for the {\em incomplete} data set case. 

Multiple instance classification requires calculations of the  probability  $P(A|\{X_i\})$. 
Because instances are independent the straightforward formula reads
\begin{eqnarray}\label{eq:pax}
	\tilde{P}(A|\{X_i\})&& = \\
	&&\frac{\prod_{i=1}^N \tilde{P}(A|X_i)}{\prod_{i=1}^N \tilde{P}(A|X_i)+\prod_{i=1 }^N\left(1 - \tilde{P}(A|X_i)\right)} \nonumber. 
\end{eqnarray}

\begin{figure}
	\begin{center}
		\includegraphics[width=\singlefigwidth]{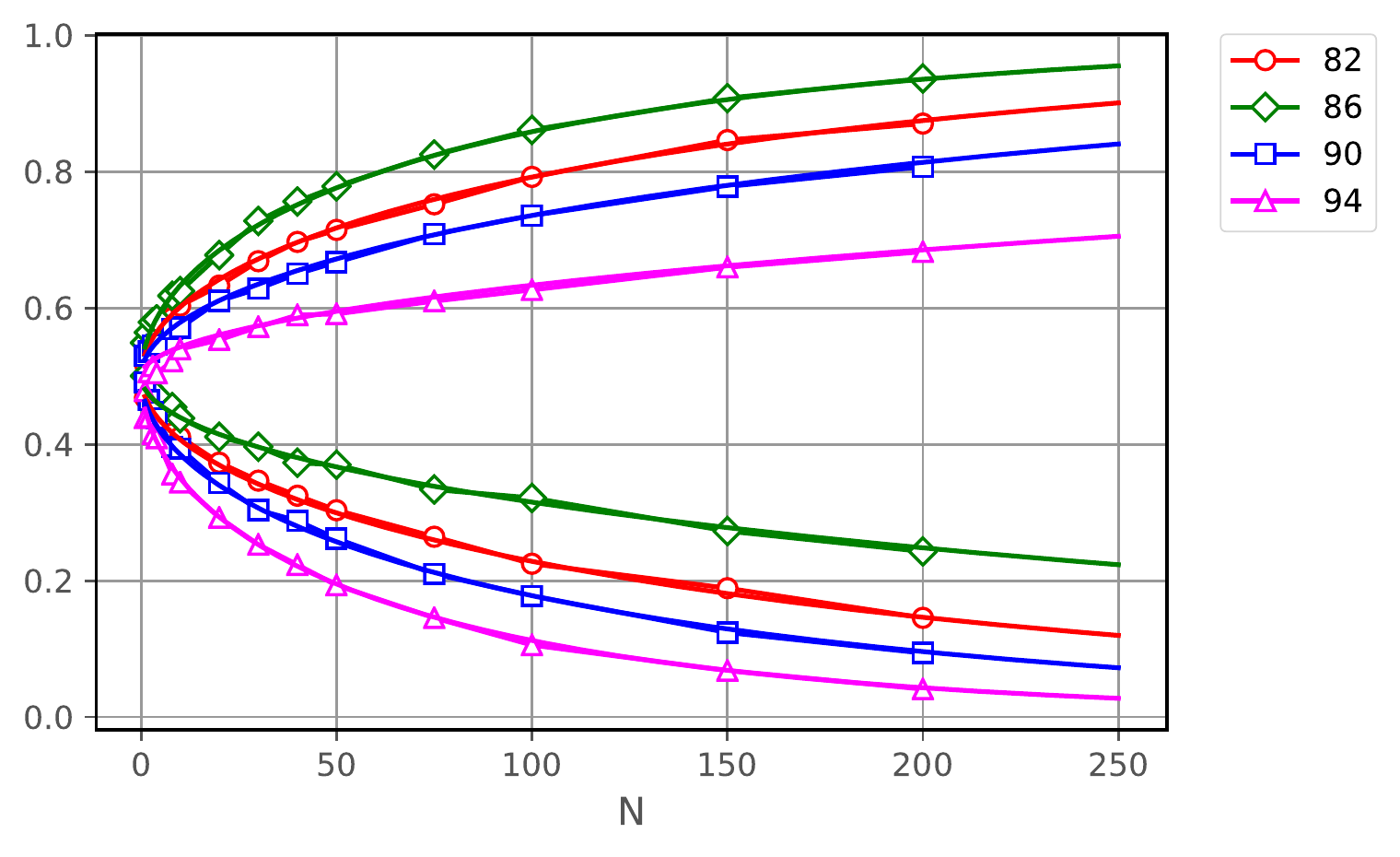}
	\end{center}
	\caption{\label{fig:multi-rates}TPR (upper prongs) and FPR (lower prongs) for the multi-instance classifier with
          fixed threshold $\theta=0.5$ and {\em incomplete} dataset.
          Different colors correspond to different number of epochs used for training DNN. Markers represent measurement
          points while the lines are representing results from formulas (\ref{eq:rates-formula}) .} 
\end{figure}

Similarly as in the single instance case, multi-instance classification consists of comparing  $\tilde{P}(A|\{X_i\})$ to some
threshold value $\theta$. In the Figure~\ref{fig:multi-rates} we show results for the true and false positive rates corresponding
to threshold $\theta=0.5$, for four different DNN trained and as a function of sample sizes $N$ used in formula~(\ref{eq:pax}). 
As already observed in the single instance classifier and fixed threshold value, both TPR and FPR vary  even between DNNs trained
with number of epochs different by one. This is of no surprise. The differences shown in Fig.~\ref{fig:multi-roc} get magnified when
multiplying probabilities in formula ~(\ref{eq:pax}). The spread of predicted TPR for sequence of 200 instances is as large as between 70\% and 90\% for the number of epochs used between 82 and 94. 
We can regularize this effect by choosing the optimal classification threshold as indicated in the previous section.

\subsection{True and false positives rate}
\label{sec:multi-rates}

Condition on the multi-instance classifier applied to sequence of $N$ instances ${X_i}$
\begin{equation}
	P(A|\{X_i\})>\theta 
\end{equation}
is equivalent to imposing condition on the single-instance classifier
\begin{equation}
	(1-\theta)\prod_{i=1}^N \tilde{P}(A|X_i) > \theta \prod_{i=1 }^N\left(1 - \tilde{P}(A|X_i)\right). 
\end{equation}
which can be rewritten as 
\begin{equation}
	\log\frac{1-\theta}{\theta}+\sum_{i=1}^n \log\frac{\tilde{P}(A|X_i)}{\left(1 - \tilde{P}(A|X_i)\right)} > 0.
\end{equation}
Denoting 
\begin{equation}
	C(\theta)=\log\frac{1-\theta}{\theta} 
\end{equation}
we can write the expression for true and false positive rates 
	\begin{align}
		TPR(\theta)& = \int\prod_i \text{d}X_i P(X_i|A)\\
		&\phantom{\int\prod_i \text{d}}\Theta\left(C(\theta)+\sum_i\log\frac{\tilde{P}(A|X_i)}{1-\tilde{P}(A|X_i)}\right),\nonumber\\
		FPR(\theta)& = \int\prod_i \text{d}X_i P(X_i|B)\\
		&\phantom{\int\prod_i \text{d}}\Theta\left(C(\theta)+\sum_i\log\frac{\tilde{P}(A|X_i)}{1-\tilde{P}(A|X_i)}\right)\nonumber. 
	\end{align}
Those formulas can be interpreted as 
\begin{align}
       	TPR(\theta) = &P(C(\theta)+\sum_{i=1}^N Q^A_i >0)\\
	 FPR(\theta) = &P(C(\theta)+\sum_{i=1}^N Q^B_i >0) 
\end{align}
where  variable $Q^A_i$ is defined as
\begin{equation}
	Q^A_i = \log\frac{\tilde{P}(A|X_i)}{1-\tilde{P}(A|X_i)}\text{ with } X_i\sim P(X_i|A) 
\end{equation}
 and $X \sim P(X)$ denotes that variable $X$ has probability density  equal to $P(X)$. Similarly 
\begin{equation}
	Q^B_i = \log\frac{\tilde{P}(A|X_i)}{1-\tilde{P}(A|X_i)}\text{ with } X_i\sim P(X_i|B) 
\end{equation}
Using the central limit theorem we obtain that asymptotically 
\begin{equation}
	\sum_{i=1}^N Q^{A(B)}_i \underset{N\rightarrow\infty}{\sim} N(N \cdot \mu_{A(B)}, \sqrt{N}\cdot \sigma_{A(B)}) 
\end{equation}
where $N(\mu,\sigma)$ denotes normal (gaussian) distribution with mean $\mu$ and standard deviation $\sigma$.
The means $\mu_{A(B)}$ and standard deviations
$\sigma_{A(B)}$ for $Q^{A(B)}_i$ can be calculated with formulas
\begin{equation}
	\mu_A = \int\text{d}X \, P(X|A) \log\frac{\tilde{P}(A|X)}{1-\tilde{P}(A|X)} 
\end{equation}
\begin{equation}
	\sigma^2_A = \int\text{d}X \, P(X|A) \left(\log\frac{\tilde{P}(A|X)}{1-\tilde{P}(A|X)}-\mu_A\right)^2 
\end{equation}
\begin{equation}
	\mu_B = \int\text{d}X \, P(X|B) \log\frac{\tilde{P}(A|X)}{1-\tilde{P}(A|X)} 
\end{equation}
\begin{equation}
	\sigma^2_B = \int\text{d}X \, P(X|B) \left(\log\frac{\tilde{P}(A|X)}{1-\tilde{P}(A|X)}-\mu_B\right)^2.
\end{equation}
Those integrals can be approximated from the data using \eqref{eq:approx-int}. 

The probability that a normally distributed variable $Q\sim N(\mu,\sigma)$ is greater than zero is 
\begin{equation}
	\begin{split}
		P(Q>0)&=\frac{1}{\sqrt{2\pi}\sigma}\int_0^\infty\!\text{d}q \,e^{-\frac{1}{2}\frac{(q-\mu)^2}{\sigma^2}}\\
		&=\frac{1}{2}\left(1+\erf\left(\frac{\mu}{\sqrt{2}\sigma}\right)\right).
	\end{split}
\end{equation}
Adding the constant to a normally distributed variable only changes its mean so finally we can estimate
\begin{eqnarray}\label{eq:rates-formula}
		TPR(\theta) \approx \frac{1}{2}\left(1+\erf\left(\frac{N\mu_A+C(\theta)}{\sqrt{2 N}\sigma_A}\right)\right)\\
		FPR(\theta) \approx \frac{1}{2}\left(1+\erf\left(\frac{N\mu_B+C(\theta)}{\sqrt{2 N}\sigma_B}\right)\right)\nonumber.
\end{eqnarray}
We have used those formula to plot the predictions for TPR and FPR as a function of $N$ and with threshold $\theta=0.5$
in the Figure~\ref{fig:multi-rates}.
The agreement between TPR and FPR estimated from formula~(\ref{fig:multi-rates}) and measured for networks trained with different
number of epochs is very good even for very low values of $N$ where the central theorem is not expected to hold.

From the above it follows that we can consider the quantities 
\begin{equation}
	\frac{\mu_A}{\sigma_A}\quad\text{and}\quad \frac{\mu_B}{\sigma_B} 
\end{equation}
as a measure of quality of the multi-istance classifier. In particular if $\mu_A>0$ and $\mu_B<0$ the classifier will converge to a
{\em perfect} classifier for large $N$ and any threshold $0<\theta<1$.

\subsection{Optimal threshold}
\begin{figure}
	\begin{center}
		\includegraphics[width=\singlefigwidth]{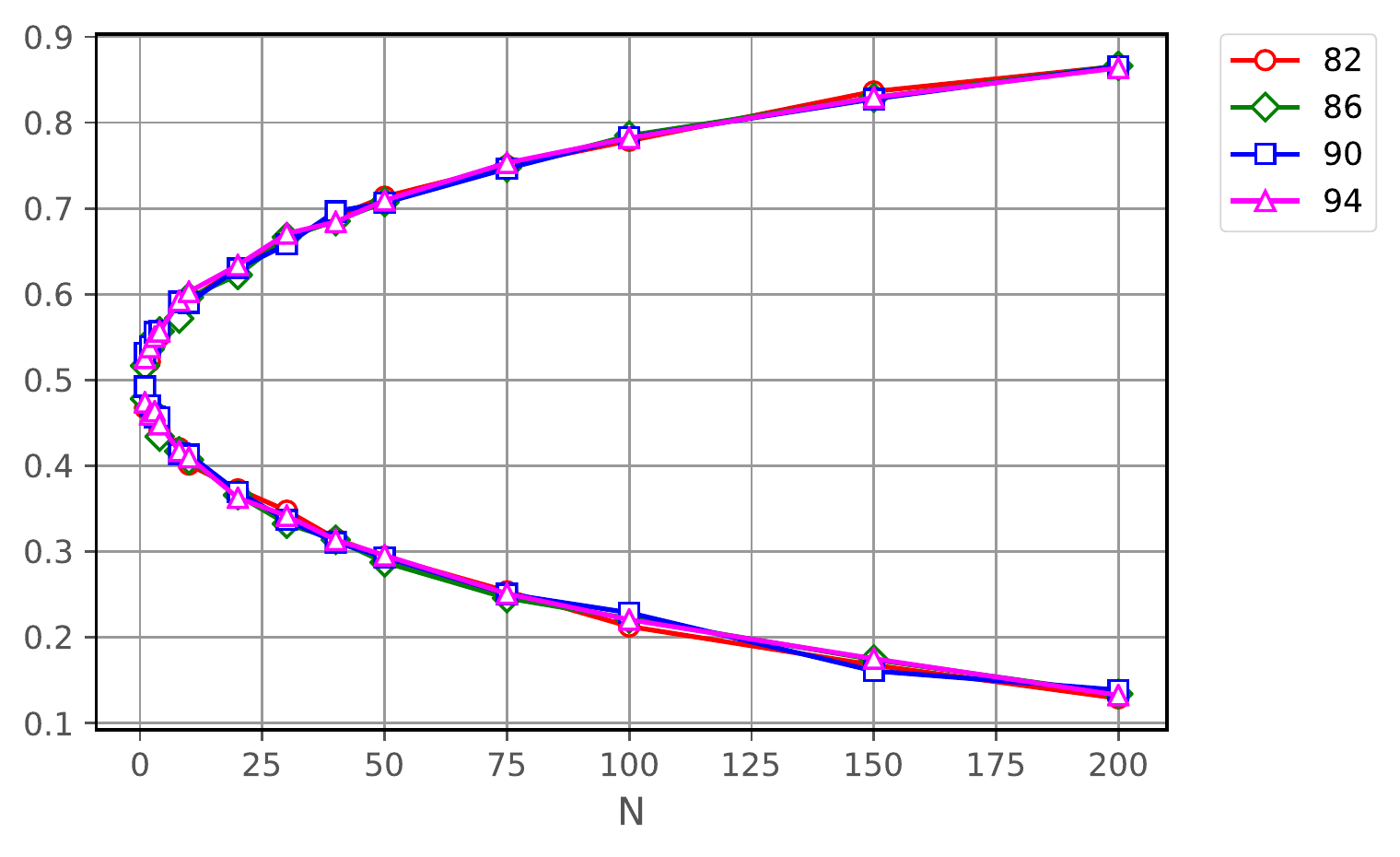}
	\end{center}
	\caption{\label{fig:multi-class-opt} TPR (upper prongs) and FPR (lower prongs) for multi-instance classifier and
         optimally adjusted threshold $C_{opt}$.}
\end{figure}

To find the optimal threshold for each network we need some criterion. We will use as such minimization of the total number of
misclassifications given by 
\begin{equation}
	MISS(\theta) = 1-TPR(\theta)+FPR(\theta) 
\end{equation}
but of course any other combination is possible. 
Inserting the formulas derived in previous section we obtain
\begin{equation}
	\begin{split}
		M&ISS(\theta) =  1+\frac{1}{2}\Biggl[\\
		&\erf\Bigg(\frac{N\mu_B+C(\theta)}{\sqrt{2 N}\sigma_B}\Biggl)-\erf\Biggl(\frac{N\mu_A+C(\theta)}{\sqrt{2 N}\sigma_A}\Biggr)
		\Biggr]
	\end{split}
\end{equation}
giving the equation for optimal $C$
\begin{equation}\label{eq:c-opt-eq}
	\begin{split}
		\frac{\text{d}MISS(\theta)}{\text{d}\theta} = &
			\frac{\text{d}C(\theta)}{\text{d}\theta} \sqrt{\frac{2}{\pi N}}\cdot\\
			&\kern-10mm\Bigl( \frac{1}{\sigma_A} e^{-\frac{(C+\mu_B N)^2}{2 N \sigma^2}} -\frac{1}{\sigma_B}e^{-\frac{(C+\mu_A N)^2}{2 N \sigma^2}} \Bigr)=0.
	\end{split}
\end{equation}
We have found out that in all cases $\sigma_A\approx\sigma_B$, so we set $\sigma=\sigma_A=\sigma_B$   which simplifies equation \eqref{eq:c-opt-eq} to
\begin{equation}
	(C_{opt}+\mu_B N)^2=(C_{opt}+\mu_A N)^2 
\end{equation}
with solution 
\begin{equation}
	C_{opt} = -\frac{1}{2}N (\mu_A+\mu_B), \qquad \theta_{opt}=\frac{1}{1+e^{C_{opt}}}.  
\end{equation}
In Figure~\ref{fig:multi-class-opt} we show the estimated TPR and FPR  using the optimal thresholds $C_{opt}$ and as expected
the results for different networks now coincide. 

\subsection{AUC score} 

In the same way we can calculate the AUC score for the multi-instance classifier.  
The formula \eqref{eq:AUC} in this case takes the form
\begin{equation}
	\begin{split}
		AUC = \int & \prod_i \text{d}X^A_i P(X^A_i|A)\prod_i \text{d}X^B_i P(X^B_i|B) \\
		&\Theta(\tilde{P}(A| \{X^A_i\}) - \tilde{P}(A| \{X^B_i\})) 
	\end{split}
\end{equation}
The condition 
\begin{equation}
	P(A| \{X^A_i\})> P(A| \{X^B_i\}) 
\end{equation}
after some manipulations can be rewritten as
\begin{equation}
	\begin{split}
	    &\sum_i\log\frac{\tp(A|X^A_i)}{(1-\tp(A|X^A_i)} \\
        &\phantom{\sum_i\log}- \sum_i\log\frac{\tp(A|X^B_i)}{(1-\tp(A|X^B_i))}>0. 
	\end{split}
\end{equation}
As described in previous section those sums can be approximated by gaussian random variables.
Their difference is also a gaussian variable so finally 
\begin{equation}
  \label{Fig:AUC}
	AUC(N) = \frac{1}{2}\left(1+\erf\left(\sqrt{N}\frac{\mu_A-\mu_B}{\sqrt{2}\sqrt{\sigma_A^2+\sigma_B^2}}\right)\right). 
\end{equation}
\begin{figure}
	\begin{center}
		\includegraphics[width=\singlefigwidth]{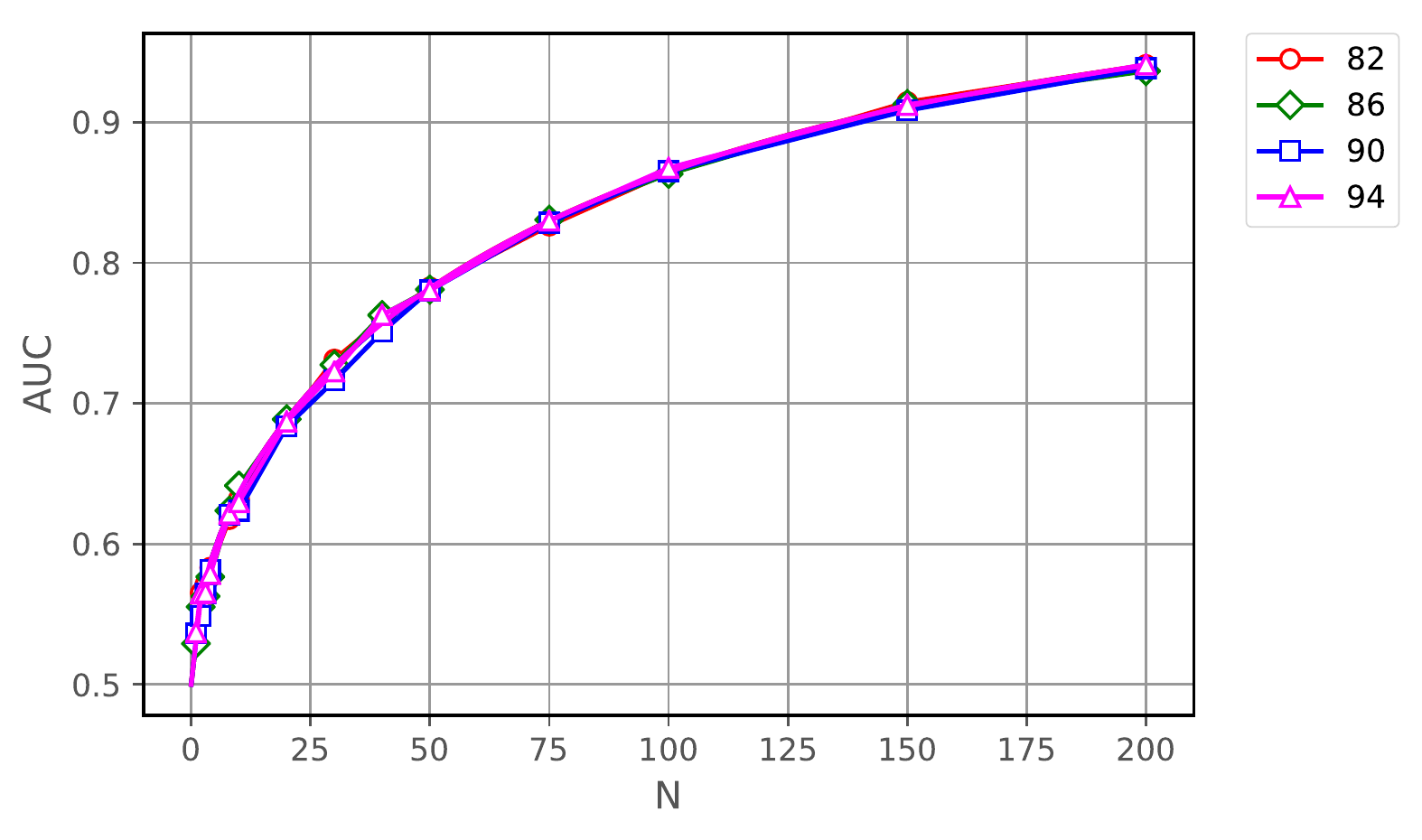}
	\end{center}
	\caption{\label{fig:multi-class-auc}AUC score for the multi-instance classifier. Shown are results from
         neural network with different number of epochs used for training and from formula (\ref{Fig:AUC}).} 
\end{figure}
As can be seen in  Figure~\ref{fig:multi-class-auc} the resulting formula is accurate down to $N=1$. 

\section{Summary and discussion}

In this paper we investigated properties of a  multi-instance classifier applied
to the measurements of the CP state of the Higgs boson in $H\rightarrow\tau\tau$ decays.
The problem was framed as binary classifier applied to individual instances.
Then the prior knowledge that the instances belong to the same class was used to define the
multi-instance classifier. Its final score was calculated as multiplication of single
instance scores for a given serie of instances. We discussed properties of such classifier and
derived formula for the optimal threshold which, when applied single classifier,
regularise (stabilise) FPR, TPR and AUC curves of the multi-instance classifier.

Taking as an example problem of measuring CP state of the Higgs boson in $H \to \tau \tau$ channel,
we have shown that for realistic scenario as considered in~\cite{Jozefowicz:2016kvz} starting from
AUC~=~0.535 for the single-instance classifier, we can reach AUC~=~0.95 for the multi-instance
classifier after analysing serie of N~=~200 instances or close to 0.85 for N~=~100.
This result is quite stable vs variation of the single-instance classifier due to eg. slight difference
in the number of epochs used for the training. This stability is achieved thanks to introducing
optimal classification threshold to the single-instance classifier.

\section*{Acknowledgements} 
D. Nemeth and E. Richter-Was were supported in part from funds of Polish 
National Science Center under decisions UMO-2014/15/B/ST2/00049 and by 
PLGrid Infrastructure of the Academic  Computer Centre CYFRONET AGH  
in Krakow, Poland, where  majority of Monte-Carlo calculations were 
performed.

\end{document}